\documentclass[
hf,
]{ceurart}

\usepackage{listings}
\usepackage[]{rotating}
\usepackage{url}
\usepackage{float}
\usepackage{subcaption}
\usepackage{todonotes}
\begin{document}

\copyrightyear{2024}
\copyrightclause{Copyright for this paper by its authors.
  Use permitted under Creative Commons License Attribution 4.0
  International (CC BY 4.0).}

\conference{LIRAI’24: 1st Legal Information Retrieval meets Artificial Intelligence Workshop  co-located with the 35th ACM Hypertext Conference, September 10, 2024, Poznan, Poland}

\title{Analyzing Bias in Swiss Federal Supreme Court Judgments Using Facebook's Holistic Bias Dataset: Implications for Language Model Training}

\author[1,2]{Sabine Wehnert}[%
orcid=0000-0002-5290-0321,
email=sabine.wehnert@ovgu.de
]

\address[1]{Otto von Guericke University Magdeburg, Germany}
\address[2]{Leibniz Institute for Educational Media | Georg Eckert Institute, Brunswick, Germany}
\cormark[1]

\author[1]{Muhammet Erta\c{s}}

\author[1,2]{Ernesto William {De Luca}}[%
orcid=0000-0003-3621-4118]
\cortext[1]{Corresponding author.}

\begin{abstract}
  Natural Language Processing (NLP) is vital for computers to process and respond accurately to human language. However, biases in training data can introduce unfairness, especially in predicting legal judgment. This study focuses on analyzing biases within the Swiss Judgment Prediction Dataset (SJP-Dataset). Our aim is to ensure unbiased factual descriptions essential for fair decision making by NLP models in legal contexts. We analyze the dataset using social bias descriptors from the Holistic Bias dataset and employ advanced NLP techniques, including attention visualization, to explore the impact of ``dispreferred'' descriptors on model predictions. The study identifies biases and examines their influence on model behavior. Challenges include dataset imbalance and token limits affecting model performance. 
\end{abstract}


\maketitle

\section{Introduction}
Natural Language Processing (NLP) is a collection of methods in Artificial Intelligence (AI) for enabling computers to model and respond accurately to human language. However, NLP models can inherit biases from training data, affecting fairness in applications such as legal judgment prediction. The Swiss Judgement Prediction Dataset (SJP Dataset), with 85,274 Swiss court cases, is a significant resource for training models to predict judicial outcomes. Our study analyzes biases within the dataset, particularly in factual case descriptions, aiming to ensure unbiased descriptions for fair decision making by NLP models in legal contexts. Our contributions are: 
\begin{itemize}
    \item We analyze the SJP Dataset based on bias descriptors from the Holistic Bias dataset.
    \item We explore how ``dispreferred'' descriptors in the dataset influence model predictions. 
\end{itemize}

In short, our goal is to identify biases in the SJP dataset and to see if they translate into model behavior, when used as training data. The remainder of this work is organized as follows: In Section \ref{sec:related}, we present related work on bias analysis. In Section \ref{sec:method}, we describe our method conceptually and evaluate the results in Section \ref{sec:eval}. We conclude the work in Section \ref{sec:conclude}.

\section{Related Work\label{sec:related}}
In the legal domain, work focuses on the decision making of judges, comparing a theoretically fair judge to an actual one regarding representation bias (e.g., offenses or arrests appear more frequently in a specific social group) and sentencing disparities (i.e., the judgment differs among social groups even if their cases are similar) \cite{DBLP:journals/corr/abs-2109-09946}. Also the decision of a judge can be influenced by gender attitudes, as shown by Ash et al. \cite{ash2024gender}. Furthermore, Ash et al. investigated whether Indian judges favor defendants who share their gender or religious identity \cite{DBLP:conf/dev/BhowmickNAASGCD21}. To the best of our knowledge, there is no related work focusing on bias in Legal Judgment Prediction tasks.

In general, Natural Language Processing (NLP) models have the ability to learn stereotypes, misrepresentations, or negative generalizations of certain social groups, genders, religions, and races. Numerous studies show that unrestricted training of natural language models can adopt social biases \cite{DBLP:conf/nips/BolukbasiCZSK16,DBLP:journals/corr/IslamBN16,DBLP:conf/naacl/GonenG19}. Research by Bolukbasi et al. \cite{DBLP:conf/nips/BolukbasiCZSK16} and Islam et al. \cite{DBLP:journals/corr/IslamBN16} demonstrates that word embeddings encode social biases related to gender roles and professions, such as associating engineers with men and nurses with women, leading downstream applications to reflect these biases. Sevim et al. also work with word embeddings to identify encoded gender biases in the legal domain \cite{DBLP:journals/nle/SevimSK23}. Similarly, Gumusel et al. \cite{DBLP:conf/iconference/GumuselMDAL22} have analyzed racial or ethnic bias with word2vec embeddings trained on legal data. 

The use of descriptors to measure social bias started as a method to specifically examine gender associations in static word embeddings \cite{DBLP:conf/nips/BolukbasiCZSK16,DBLP:journals/corr/IslamBN16}. Since contextual word embeddings consider context, templates were necessary to measure social bias, like stereotypical associations with other text content \cite{DBLP:conf/nips/TanC19}. Various studies have proposed templates to measure bias \cite{DBLP:conf/naacl/RudingerNLD18,DBLP:conf/nips/TanC19}, some selecting sentences from texts and heuristically swapping demographic terms \cite{DBLP:conf/emnlp/DinanFWUKW20,DBLP:conf/naacl/ZhaoWYCOC19} or using machine learning systems for descriptor replacement \cite{DBLP:conf/emnlp/QianRFSKW22}. The presence of descriptors is not sufficient to create bias in a language model. They need to be significantly associated with one particular label, such that a bias is likely to be learned.
Therefore, we \cite{DBLP:conf/jsai/WehnertMNL24} employed the binomial significance test for detecting language artifacts - another type of bias - in the datasets of the Competition on Legal Information Extraction/Entailment (COLIEE). While there are  benchmark datasets for social bias like SEAT \cite{DBLP:conf/naacl/MayWBBR19}, StereoSet~\cite{DBLP:conf/acl/NadeemBR20}, and Holistic Bias \cite{DBLP:conf/emnlp/SmithHKPW22}, this study focuses on bias analysis in the legal domain using the Swiss Judgement Prediction (SJP) Dataset. Previously, we worked on this dataset regarding sentiment and subjectivity bias \cite{DBLP:conf/lirai/WehnertPL23}. The SJP Dataset \cite{DBLP:conf/acl-nllp/NiklausCS21} is a dataset based on multilingual court judgment facts, offering annotations on whether a judgment is positive (approval) or negative (dismissal). The Swiss court judgments are written in three languages: approximately 50,000 in German, 31,000 in French, and 4,000 in Italian, with a total of 3/4 of them being dismissed judgments. While previous research has focused on gender bias in datasets and models, this study conducts a bias investigation (e.g., religion, race, nationality) in Legal Judgment Prediction using the Holistic Bias Dataset~\cite{DBLP:conf/emnlp/SmithHKPW22}, which covers 13 demographic axes, over 600 descriptors, and 26 templates.

\section{Bias Analysis in Swiss Federal Court Judgments}\label{sec:method}
In this section, the process is explained, from analyzing the SJP Dataset for bias using descriptors, to measuring the impact on model performance and potentially obtained biases.
\begin{figure}
  \centering
  \includegraphics[width=\linewidth]{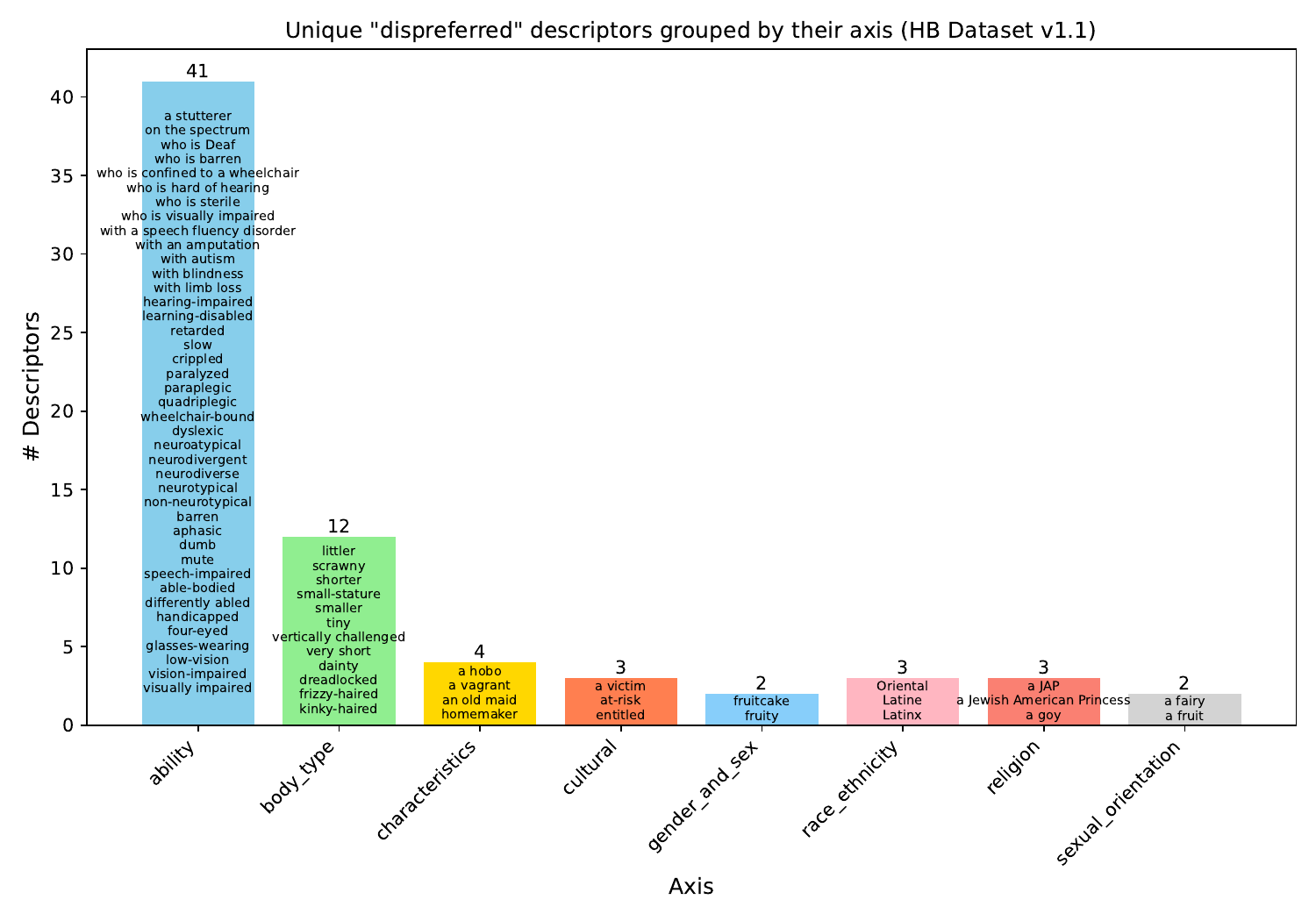}
  \caption{Holistic Bias Dataset - Version 1.1 with respectively “dispreferred” labeled descriptors.}
\label{fig:descriptors}
\end{figure}
\subsection{Selecting Bias-Descriptors}
For the bias analysis in the facts of court rulings in the Swiss Judgement Prediction Dataset (SJP Dataset) \cite{DBLP:conf/acl-nllp/NiklausCS21}, the descriptors labeled as “dispreferred” from the Holistic Bias Dataset \cite{DBLP:conf/emnlp/SmithHKPW22} are used. The Holistic Bias Dataset includes two versions. The original Version 1.0 contains 620 unique descriptors, 48 of which are labeled (by experts) as “dispreferred” and are distributed across 3 different demographic axes. In contrast, the new Version 1.1 includes 769 unique descriptors, with 70 labeled as “dispreferred” and distributed across 8 different demographic axes, as depicted in Figure \ref{fig:descriptors}. The new version is used in this work.

\subsection{Dataset Translation}\label{subsec:translate}
\begin{figure}
  \centering
  \includegraphics[width=0.9\linewidth]{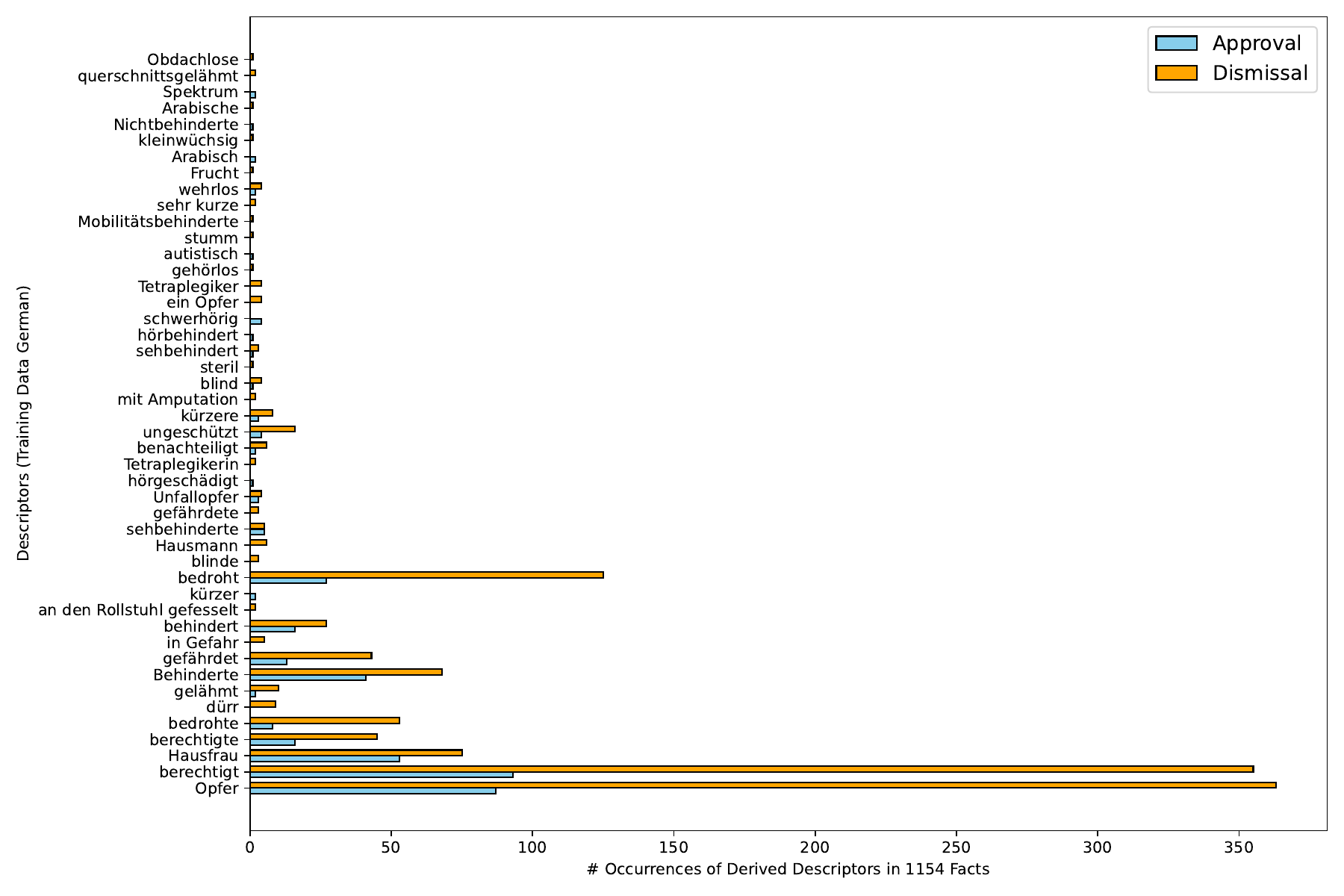}
  \caption{Occurrences of Derived Descriptors in German Training Data, by Instance Label.}
\label{fig:descriptors_labels}
\end{figure}

We automatically translate the descriptors that were originally recorded in English in the Holistic Bias Dataset into German, French, and Italian using the DeepL API\footnote{\url{https://www.deepl.com/en/pro-api?cta=header-pro-api}}, as the SJP Dataset's court rulings are written in these languages. After translation, a manual expansion and derivation of synonyms, plural, and gender forms are performed using Google Translate\footnote{\url{https://translate.google.com/?}}. This increases the number of descriptors and allows for a more comprehensive analysis of undesirable terms. Gender forms are intentionally used to detect potential double biases. Since there was the translation step and additional similar descriptors have been derived from the descriptors originally labeled as “dispreferred”, the term “derived dispreferred” descriptors is used instead of “labeled” in the following. Figure \ref{fig:descriptors_labels} depicts the distribution of derived descriptors and their labels for the German training data, as an example. Note that we do not count training instances in our figures, but descriptors instead. However, most training instances contain up to two descriptors. We can see several descriptors occurring overly frequent with the label ``dismissal'', in particular ``berechtigt'' and ``Opfer''. This is not surprising, as these are regular words in legal contexts, but the predominant co-occurrence with ``dismissal'' is of interest.

\subsection{Preprocessing}
For preprocessing, we employ two approaches to preserve the original text content of the facts from the SJP Dataset while adhering to the 512-token limit of the language model\footnote{\url{https://huggingface.co/joelniklaus/legal-swiss-roberta-large}} which is be used for judgment prediction.
\paragraph{Extractive Summarization: }The first approach involves using the \texttt{LexRank Summarizer}\footnote{\url{https://github.com/miso-belica/sumy}} to condense the text content of the facts, ensuring the 512-token limit of the model is not exceeded. LexRank calculates sentence similarity using cosine similarity and creates a graph structure with sentences as nodes and similarities as edges. The PageRank algorithm then identifies the key sentences that represent the main content, which are combined into a summary containing the core information. In this work, the goal was to keep the summarized text under 512 tokens. Since the LexRank Summarizer in \texttt{sumy} takes only the text to be summarized and the number of sentences as input, a workaround was used. Each text was summarized into 3 to 26 sentences, and the token count was checked. If the summary exceeded 512 tokens, the last added sentence was removed to meet the limit.
\paragraph{Chunking: } This approach aims to retain the entire text content by splitting facts longer than 510 tokens into chunks. Each chunk's text length was limited to 300 words, as this approximates 500-520 tokens, fitting within the model's token limit. Though some chunks exceeded 510 tokens due to varying word lengths, their number was minimal (94 of 36,386 test chunks and 158 of 123,207 training chunks), so the 300-word limit was kept. This ensures each chunk utilizes the token limit efficiently, avoiding many small chunks. During tokenization, chunks were truncated to 512 tokens. From 59,709 training data points, 123,207 chunks were created, and from 17,357 test data points, 36,386 chunks were created.

\subsection{Model Fine-Tuning}
We proceed with the fine-tuning of models using the \texttt{legal-swiss-roberta-large} pretrained model from Niklaus et al. \cite{DBLP:journals/corr/abs-2306-02069}, which covers 24 languages and is specifically trained on legal sources. Parameters such as learning rate (\textit{2e-5}), seed values, batch size (20 per GPU, equals 40 for each training step), and weight decay (\textit{0.01}) are configured for training on the Swiss Judgement Prediction (SJP) Dataset. Due to the dataset's imbalance (76.23\% ``dismissal'' and 23.77\% ``approval''), we adjust class weights to mitigate bias towards the majority class during fine-tuning. Three models per approach are fine-tuned with different seeds and uploaded to the Hugging Face Hub\footnote{\url{https://huggingface.co/collections/mhmmterts/}}, followed by performance analysis on test data to evaluate precision, recall, accuracy, and F1-score.

\subsection{Dataset Analysis with the Binomial Significance Test}
When analyzing ``dispreferred'' derived descriptors in the facts of the training data, it is important to evaluate if these descriptors are already biased towards a specific class, i.e., ``dismissal'' or ``approval''. We use the Binomial Significance Test (BST) to determine whether the observed frequency of a binary outcome (dismissal=\(0\) and approval=\(1\)) significantly deviates from a hypothesized probability distribution. We conduct two separate BSTs: one for the ``dismissal'' outcome and another for the ``approval'' outcome. Below, we explain BST and introduce the key concepts and variables relevant to our dataset.
\subsubsection{Overview of the Dataset and Null Hypothesis}
The training dataset consists of 59,709 data points, of which 45,516 correspond to the ``dismissal'' outcome. This yields a relative frequency of dismissal of approximately  $45516 / 59709 \approx 0.7623$.  We use this value as our estimate for the probability of a dismissal outcome, $\pi_0=0.762$, while the corresponding probability of an approval outcome is  0.2377. The null hypothesis $H_0$ for the ``dismissal'' outcome is defined as: $H_0:\pi_0=0.762$.
Assuming this is the true distribution, the probability of observing $k$ dismissals out of $n$ data points is given by the binomial formula:
\[ P(X=k)=\binom{n}{k}\pi_0^k(1-\pi_0)^k\]
\subsubsection{Testing the ``Dismissal'' Outcome for a Specific Token}
To determine whether the ``dismissal'' outcome for the token ``victime'' deviates significantly from the expected behavior under $H_0$, we analyze its observed frequencies (see also Table \ref{tab:binomial_test_result_0_chunked}). The token ``victime'' appears 3,928 times, with 3,132 occurrences labeled as ``dismissal''. Under 
$H_0$, the expected number of dismissals is: \[\text{Expected count}= 3928\cdot \pi_0\approx 2993 \]
The observed count (3,132) exceeds the expected count (2,993). To evaluate whether this deviation is statistically significant, we calculate the probability of observing 3,132 or more dismissals under $H_0$, also known as the p-value:
\[\text{p-value}=\sum_{k=2993}^{3928}\binom{n}{k}\pi_0^k(1-\pi_0)^k=8.363595\cdot 10^{-8} \]
That p-value in the example is extremely unlikely, if one assumes $H_0$. In fact, we can consider every event with probability smaller than the significance threshold of $0.1$ as improbable under the null hypothesis, which we explain in the following. 
\subsubsection{Significance Threshold}

We adopt a significance level
(\(\alpha\)) of 0.1, meaning that any event with a p-value smaller than 0.1 is considered improbable enough to reject $H_0$.
Since the computed p-value is much smaller than 
$\alpha$, we reject $H_0$ for the dismissal outcome of the token ``victime''.
This result suggests that the observed frequency of dismissals for this token significantly deviates from the expected distribution, indicating a potential bias.

\subsubsection{Testing Both Outcomes}
We perform similar tests for each descriptor on both labels dismissal=\(0\) ($H_0:\pi_0=0.762$) and approval=\(1\) ($H_0:\pi_0=
0.2377$) with their respective counts. A p-value below the significance level suggests that the observed frequencies of dismissal=\(0\) and approval=\(1\) are unlikely under the null hypothesis, indicating a significant deviation. 
This statistical test compares observed frequencies of descriptors in the data against expected ones, helping identify whether certain descriptors may influence  predictions in classifications when a model is trained on this potentially biased data.

\subsection{Analysis of Language Model Performance}
The results of the performance analysis of models fine-tuned with different seeds and with different preprocessing strategies are listed in the following Tables \ref{tab:performance_summarized_model_16}-\ref{tab:performance_chunked_model_48}. All models exhibit a higher F1-score for the label ``dismissal'' than for ``approval''. During the performance analysis of the model tested with facts divided into chunks, the final classification of a fact is determined based on the majority of its predicted chunk results. In the special case that the predicted chunks are evenly split between two classes, the class ``dismissal'' is chosen as the final classification result. This decision is made due to the imbalanced labels in the training and test sets of the SJP Dataset, where approximately 80\% of the labels correspond to ``dismissal'', aiming to achieve higher prediction accuracy for final results. Another challenge encountered during fine-tuning the model with chunked facts was its tendency, even after using class weighting, to favor the majority class. Therefore, the first successfully fine-tuned model that provided balanced predictions was selected for further analysis of predicted facts with descriptors. Using seed 48, balanced classification results were achieved. The results of the chunking strategy are shown in Table \ref{tab:performance_chunked_model_48}. Although it is important to obtain good model performance, this is not the main focus of this investigation. Instead, in the evaluation section we show results on the biases picked up by the language models. For this, we compare the correct/wrong predictions to the respective class label and see if a tendency from the dataset bias was adopted by the model. Furthermore, we analyze model attention attribution profiles using the \texttt{transformers-interpret}\footnote{\url{https://github.com/cdpierse/transformers-interpret}} library. Technically, attention visualization calculates word attributions of a model, showing the importance of each token within a specific context. This process highlights the weighting of words contributing to classification.
The process starts by extracting text content from the test data in the SJP-Dataset. The text is tokenized using the model's tokenizer, which includes necessary transformations, such as adding padding tokens and limiting to a maximum length of 512 tokens. Attention visualizations are conducted separately for both summarized and chunked predicted facts with descriptors. Word attribution values range from 1 to -1. A word attribution close to 1 indicates high attention and significant influence on model prediction. Conversely, a word attribution near -1 suggests low attention and minimal impact on model prediction.

\begin{table}[htbp]
    \centering
    \begin{tabular}{>{\centering\arraybackslash}p{0.15\textwidth}>{\centering\arraybackslash}p{0.15\textwidth}>{\centering\arraybackslash}p{0.14\textwidth}*{3}{>{\raggedleft\arraybackslash}p{0.13\textwidth}}}
        \toprule
        \textbf{Seed} & \textbf{Label} & \textbf{Precision} & \textbf{Recall} & \textbf{F1-Score} & \textbf{Support} \\
        \midrule
        16 & Dismissal (0) & 0.91 & 0.70 & 0.79 & 14.026 \\
           & Approval (1)  & 0.36 & 0.72 & 0.48 & 3.331 \\
        \midrule
        \raggedright\textbf{\small Accuracy} & - & - & - & 0.70 & 17.357 \\
        \raggedright\textbf{\small Macro Avg} & - & 0.63 & 0.71 & 0.63 & 17.357 \\
        \raggedright\textbf{\small Weighted Avg} & - & 0.81 & 0.70 & 0.73 & 17.357 \\
        \bottomrule
    \end{tabular}
    \caption{Successfully Trained Model\protect\footnotemark\ (Seed 16) after Extractive Summarization.}
    \label{tab:performance_summarized_model_16}
\end{table}

\footnotetext{Here, ``successful'' means that the model, when using class weighting, no longer automatically selects the majority class ``dismissal'' for classifying the facts.}

\begin{table}[htbp]
    \centering
    \begin{tabular}{>{\centering\arraybackslash}p{0.15\textwidth}>{\centering\arraybackslash}p{0.15\textwidth}>{\centering\arraybackslash}p{0.14\textwidth}*{3}{>{\raggedleft\arraybackslash}p{0.13\textwidth}}}
        \toprule
        \textbf{Seed} & \textbf{Label} & \textbf{Precision} & \textbf{Recall} & \textbf{F1-Score} & \textbf{Support} \\
        48 & Dismissal (0) & 0.90 & 0.82 & 0.86 & 14.026 \\
           & Approval (1)  & 0.45 & 0.61 & 0.52 & 3.331 \\
        \midrule
        \raggedright\textbf{\small Accuracy} & - & - & - & 0.78 & 17.357 \\
        \raggedright\textbf{\small Macro Avg} & - & 0.67 & 0.72 & 0.69 & 17.357 \\
        \raggedright\textbf{\small Weighted Avg} & - & 0.81 & 0.78 & 0.93 & 17.357 \\
        \bottomrule
    \end{tabular}
    \caption{Successfully Trained Model (Seed 48) after Extractive Summarization.}
    \label{tab:performance_summarized_model_48}
\end{table}

\begin{table}[htbp]
    \centering
    \begin{tabular}{>{\centering\arraybackslash}p{0.15\textwidth}>{\centering\arraybackslash}p{0.15\textwidth}>{\centering\arraybackslash}p{0.14\textwidth}*{3}{>{\raggedleft\arraybackslash}p{0.13\textwidth}}}
        \toprule
        \textbf{Seed} & \textbf{Label} & \textbf{Precision} & \textbf{Recall} & \textbf{F1-Score} & \textbf{Support} \\
        80 & Dismissal (0) & 0.90 & 0.86 & 0.88 & 14.026 \\
           & Approval (1)  & 0.49 & 0.58 & 0.53 & 3.331 \\
        \midrule
        \raggedright\textbf{\small Accuracy} & - & - & - & 0.80 & 17.357 \\
        \raggedright\textbf{\small Macro Avg} & - & 0.69 & 0.72 & 0.70 & 17.357 \\
        \raggedright\textbf{\small Weighted Avg} & - & 0.82 & 0.80 & 0.81 & 17.357 \\
        \bottomrule
    \end{tabular}
    \caption{Successfully Trained Model (Seed 80) after Extractive Summarization.}
    \label{tab:performance_summarized_model_80}
\end{table}

\begin{table}[h]
    \centering
    \begin{tabular}{>{\centering\arraybackslash}p{0.15\textwidth}>{\centering\arraybackslash}p{0.15\textwidth}>{\centering\arraybackslash}p{0.14\textwidth}*{3}{>{\raggedleft\arraybackslash}p{0.13\textwidth}}}
        \toprule
        \textbf{Seed} & \textbf{Label} & \textbf{Precision} & \textbf{Recall} & \textbf{F1-Score} & \textbf{Support} \\
        \midrule
        48 & \raggedright Dismissal (0) & 0.88 & 0.88 & 0.88 & 14.026 \\
           & \raggedright Approval (1)  & 0.49 & 0.48 & 0.49 & 3.331 \\
        \midrule
        \raggedright\textbf{Accuracy} & - & - & - & 0.81 & 17.357 \\
        \raggedright\textbf{Macro Avg} & - & 0.69 & 0.68 & 0.69 & 17.357 \\
        \raggedright\textbf{\small Weighted Avg} & -  & 0.81 & 0.80 & 0.81 & 17.357 \\
        \bottomrule
    \end{tabular}
    \caption{Successfully Trained Model (Seed 48) after Chunk Preprocessing.}
    \label{tab:performance_chunked_model_48}
\end{table}

\section{Evaluation}\label{sec:eval}
To evaluate the bias in the SJP Dataset, we first analyze the dataset and then its impact on the model performance, which can be reproduced via our published code\footnote{\url{https://github.com/anybass/FSCS-bias}}.
\begin{table}[htbp]
\centering
\caption{Top 5 ``dispreferred'' Descriptors which are Biased towards ``dismissal'' from the Chunked Training Data and their Probability.}
\label{tab:binomial_test_result_0_chunked}
\begin{tabular}{@{}llllllll@{}}
\toprule
 Token & Total\_Count & 0\_Count & 0\_Prob & 1\_Count & 1\_Prob&0\_P\_Value&1\_P\_Value \\ \midrule
 victime & 3928 & 3132 & 0.797352 & 796 & 0.202648 &8.363595e-08 & 1.000000\\
 berechtigt & 712 & 568 & 0.797753 & 144 & 0.202247 & 1.356222e-02 & 0.989308 \\
 intitulé & 709 & 593 & 0.836389 & 116 & 0.163611& 8.972588e-07 & 0.999999 \\
 Opfer & 672 & 535 & 0.796131 & 137 & 0.203869 & 2.061203e-02 & 0.983627 \\
 menacé	& 366 &	300 &	0.819672 &	66	& 0.180328 & 4.899652e-03 & 0.996675\\
 \bottomrule
\end{tabular}
\end{table}

\begin{table}[htbp]
\centering
\caption{Top 5 ``dispreferred'' Descriptors which are Biased towards ``approval'' from the Chunked Training Data and their Probability.}
\label{tab:binomial_test_result_1_chunked}
\begin{tabular}{@{}llllllll@{}}
\toprule
 Token & Total\_Count & 0\_Count & 0\_Prob & 1\_Count & 1\_Prob&0\_P\_Value&1\_P\_Value \\ \midrule
 en danger & 695 & 481 & 0.692086 & 214 & 0.307914 & 0.999990 & 1.422505e-05\\
Hausfrau & 144 & 81 & 0.562500 & 63 & 0.437500 & 1.000000 & 1.094425e-07\\
Behinderte & 143 & 98 & 0.685315 & 45 & 0.314685 & 0.986153 & 2.185340e-02\\
behindert & 61 & 41 & 0.672131 & 20 & 0.327869 & 0.960572 & 6.981606e-02\\
délicate & 40 & 23 & 0.575000 & 17 & 0.425000 & 0.997408 & 6.852460e-03\\
 \bottomrule
\end{tabular}
\end{table}

\begin{figure}[!htbp]
    \centering
     \small 
    \begin{subfigure}[b]{0.45\linewidth}
        \centering
        \includegraphics[width=\linewidth]{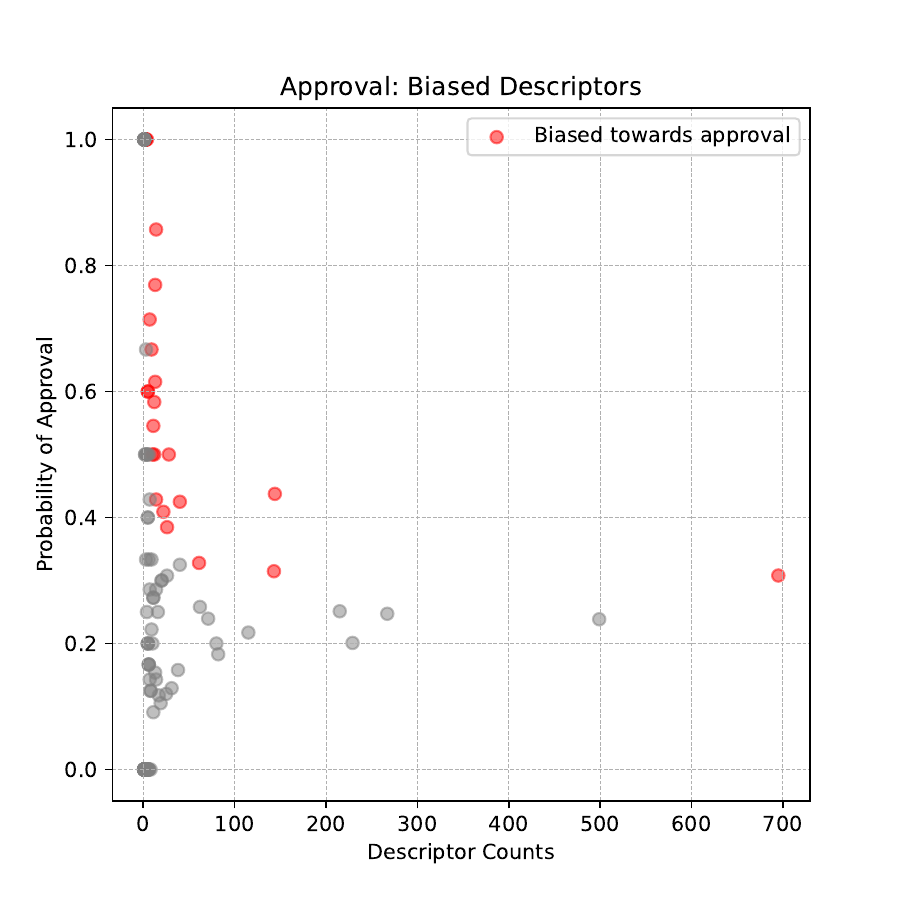}
        \label{fig:app}
    \end{subfigure}
    \hfill
    \begin{subfigure}[b]{0.45\linewidth}
        \centering
        \includegraphics[width=\linewidth]{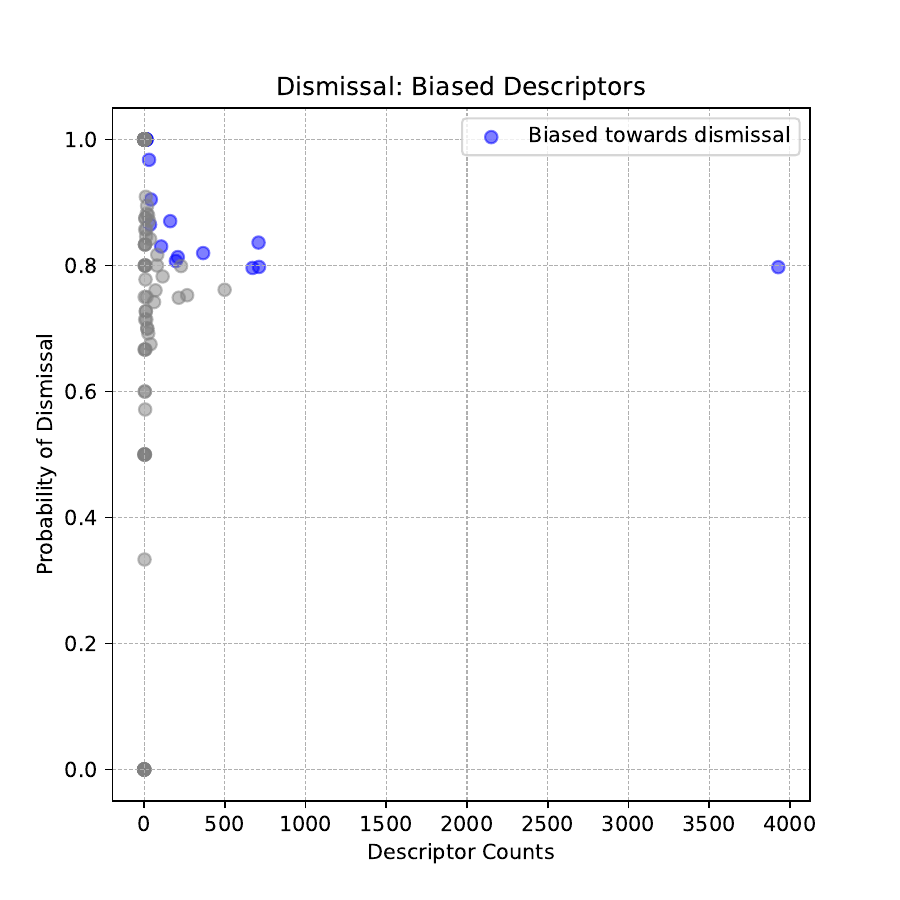}
        \label{fig:dis}
    \end{subfigure}
    \caption{Results of the binomial significance test for chunked data.}
    \label{fig:scatter_plot_chunked}
\end{figure}
\subsection{Bias in the Dataset}
The Binomial Significance Test was conducted on both, the summarized and the chunked training data. For simplicity, we only show the results on the chunked data, as there are more data points. Nevertheless, the test results of both representations align. The five most frequent descriptors are shown in Table \ref{tab:binomial_test_result_0_chunked} for dismissal and Table \ref{tab:binomial_test_result_1_chunked} for approval. We also added the p\_values for both labels, which have been calculated in the Binomial Tests. We further illustrate the results of the Binomial Significance Test in Figure \ref{fig:scatter_plot_chunked}. 

In the plot for the label ``approved'' the biased descriptor ``en danger'' occurs most frequently with a total count of 695, while for ``dismissal'' the biased descriptor ``victime'' has a potential influence due to 3928 occurences. The biased labels of the derived descriptors are listed in Table~\ref{tab:table_biased_towards_chunked}. Similar to the German training data (see Section \ref{subsec:translate}), translations of ``victim'' and ``entitled'' seem to be consistently biased towards ``dismissal'', regardless of language.

\begin{table}[htbp]
\caption{Biased ``dispreferred'' Descriptors in Chunked Training Data via Binomial Significance Test.}
\label{tab:table_biased_towards_chunked}
\centering
\begin{tabular}{|p{0.45\linewidth}|p{0.45\linewidth}|}
\hline
\multicolumn{1}{|c|}{\textbf{Label ``dismissal''}} & \multicolumn{1}{c|}{\textbf{Label ``approval''}} \\
\hline
victime & en danger \\
\hline
intitulé & Hausfrau \\
\hline
berechtigt & Behinderte \\
\hline
Opfer & behindert \\
\hline
menacé & délicate \\
\hline
bedroht & idoneo \\
\hline
autorizzato & non protetti \\
\hline
menacée & latin \\
\hline
gefährdet & ungeschützt \\
\hline
casalinga & handicapée \\
\hline
in pericolo & tétraplégique \\
\hline
arabe & trés court \\
\hline
autorizzate & délicat \\
\hline
legittima & silencieux \\
\hline
a rischio & latine \\
\hline
non protégée & sehbehinderte \\
\hline
muette & JAP \\
\hline
paralysée & schwerhörig \\
\hline
dürr & paralysé \\
\hline
maigre & cieco \\
\hline
 & Spektrum \\
\hline
 & autistisch \\
\hline
 & Arabisch \\
\hline
 & handicapé de la vue \\
\hline
 & kürzer \\
\hline
 & Latino \\
\hline
& non protégé \\
\hline
\multicolumn{1}{|c|}{\textbf{20}} & \multicolumn{1}{c|}{\textbf{27}} \\
\hline
\end{tabular}

\end{table}

\begin{figure}[h]
  \centering
  \includegraphics[width=0.9\linewidth]{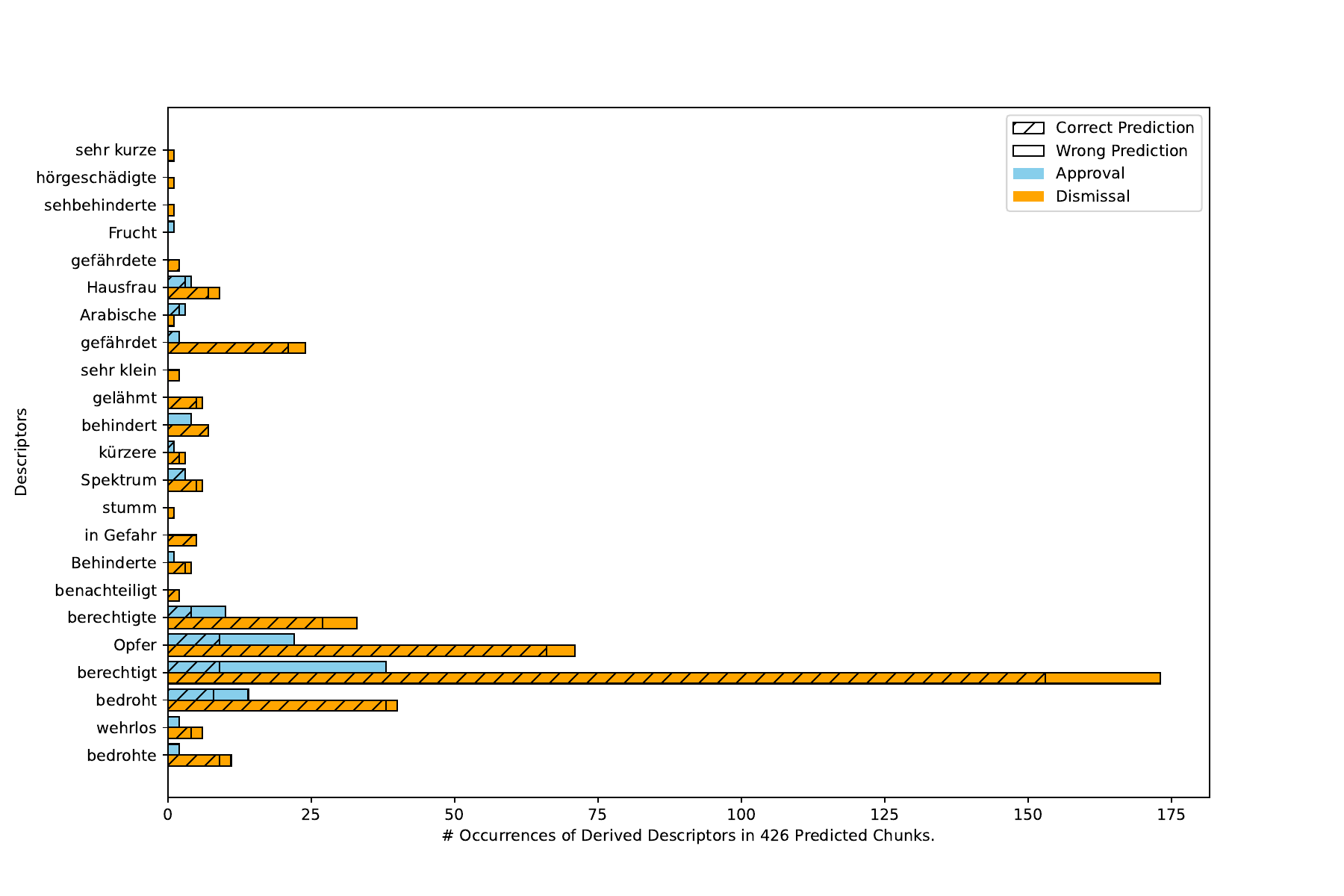}
  \caption{Classification Performance on German Test Data per Contained Descriptor.}
\label{fig:ger_performance}
\end{figure}

\begin{figure}[h]
  \centering
  \includegraphics[width=0.9\linewidth]{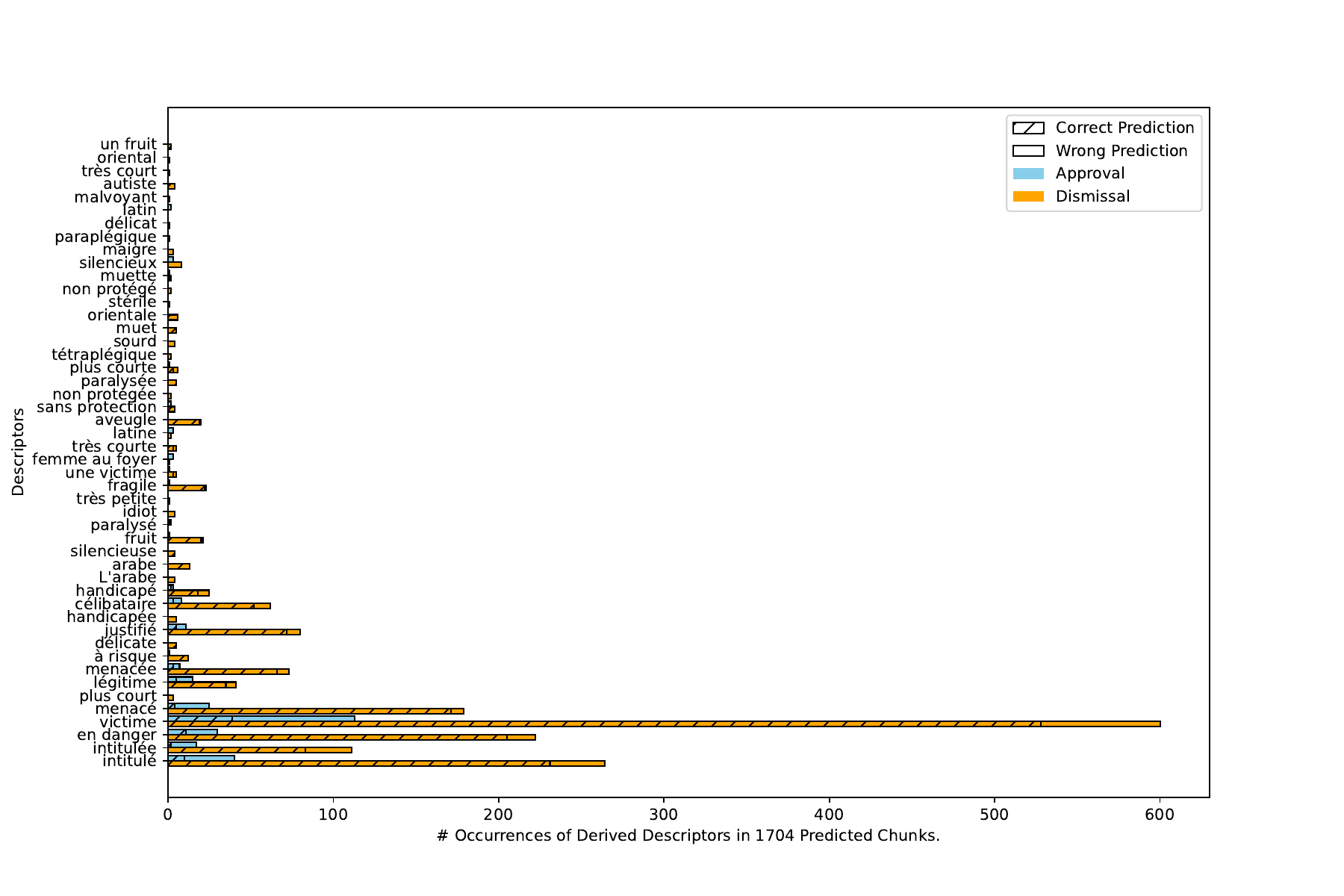}
  \caption{Classification Performance on French Test Data per Contained Descriptor.}
\label{fig:fr_performance}
\end{figure}

\begin{figure}[h]
  \centering
  \includegraphics[width=0.9\linewidth]{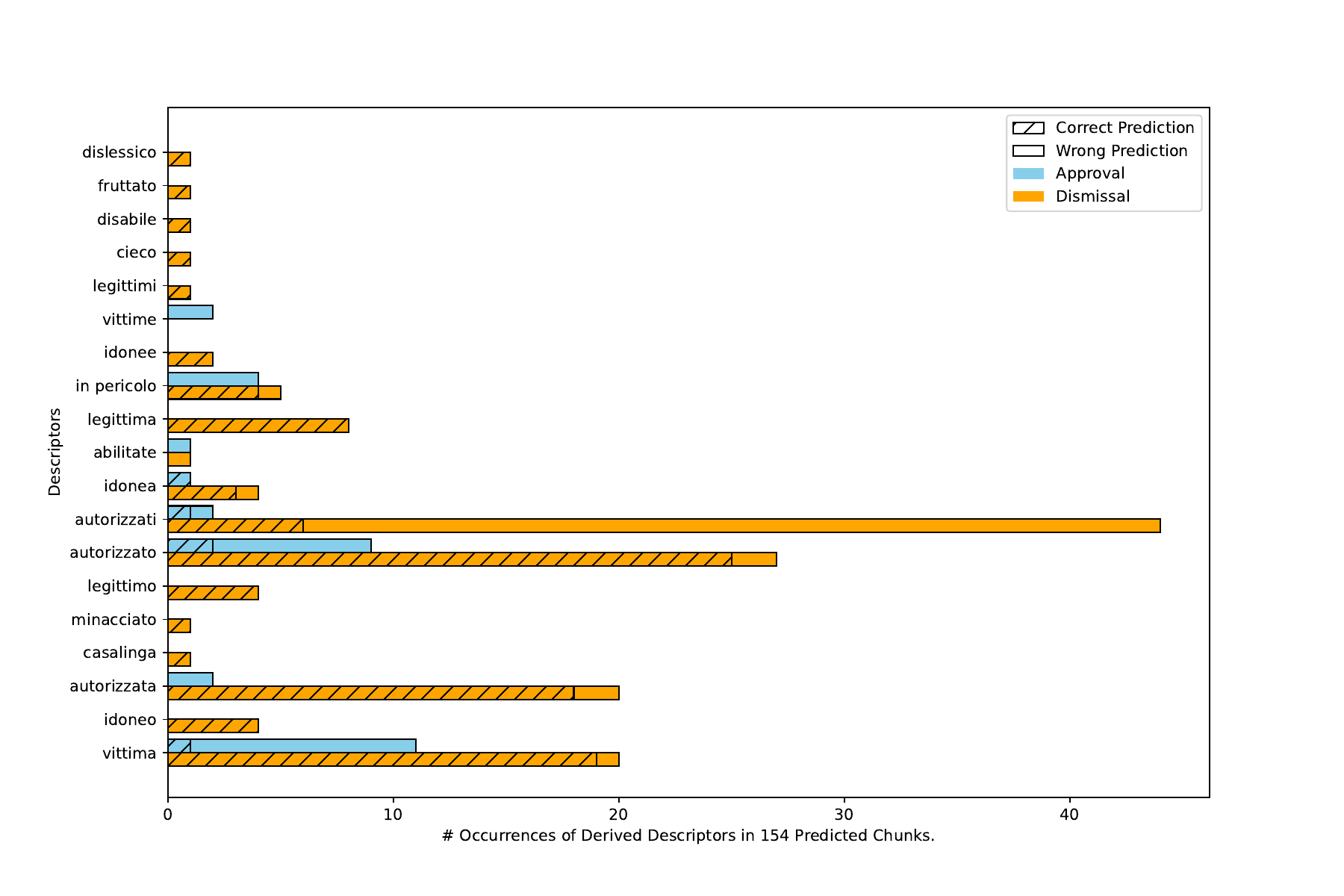}
  \caption{Classification Performance on Italian Test Data per Contained Descriptor.}
\label{fig:it_performance}
\end{figure}

\subsection{Bias in the Language Model}
In our analysis, we examined the 313 summarized facts for descriptors that have their word attributions within the top 20, 50, and 100 tokens. From the results of the descriptors with high word attributions within the TOP 50, we report that the descriptors ``bedroht'' (threatened), ``Opfer'' (victim), ``intitulé'' (entitled), ``justifié'' (justified), and ``victime'' (victim) can have a significant influence on the model's incorrect classification predictions. The descriptors ``bedroht',' ``Opfer'', ``intitulé'', and ``victime'' were identified as biased towards the class ``dismissal'' in the binomial test. Attention visualizations of the incorrect class predictions confirm that facts with these descriptors, which actually belong to the class ``approval'', are classified as ``dismissal''. For space reasons, illustrations of these results are not included in this work.

Subsequently, we examined the 508 chunks for descriptors that have their word attributions within the top 10, 20, and 30 tokens. From the results of the descriptors with high word attributions within the TOP 20, it can be seen on Figures \ref{fig:ger_performance},\ref{fig:fr_performance}, and \ref{fig:it_performance} that the descriptors ``berechtigt''~(entitled), ``Opfer'' (victim), ``bedroht'' (threatened), ``victime'' (victim),  ``intitulé'' (entitled), and ``menacé'' (threatened)  can have a significant influence on the model's classification predictions, while Italian descriptors were so low in count that their influence is questionable, despite many misclassifications in the descriptor ``autorizzati'' that was not included in the list of biased descriptors.

\begin{sidewaysfigure}[h]
  \centering
   \includegraphics[width=\textheight, height=6cm]{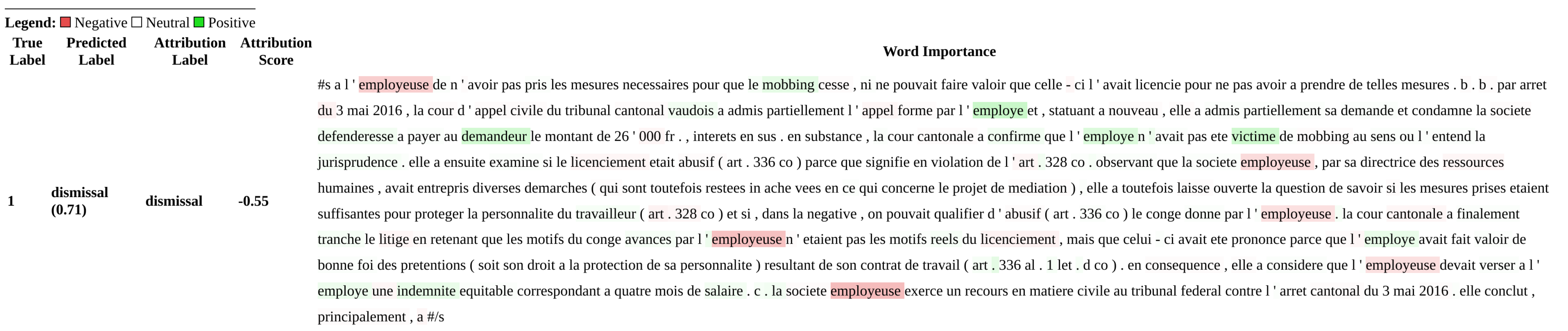}
    \caption{Attention Visualization, Chunk of a Fact, Descriptor: ``victime''}
  \label{fig:french_av}
    \includegraphics[width=\textheight, height=6cm]{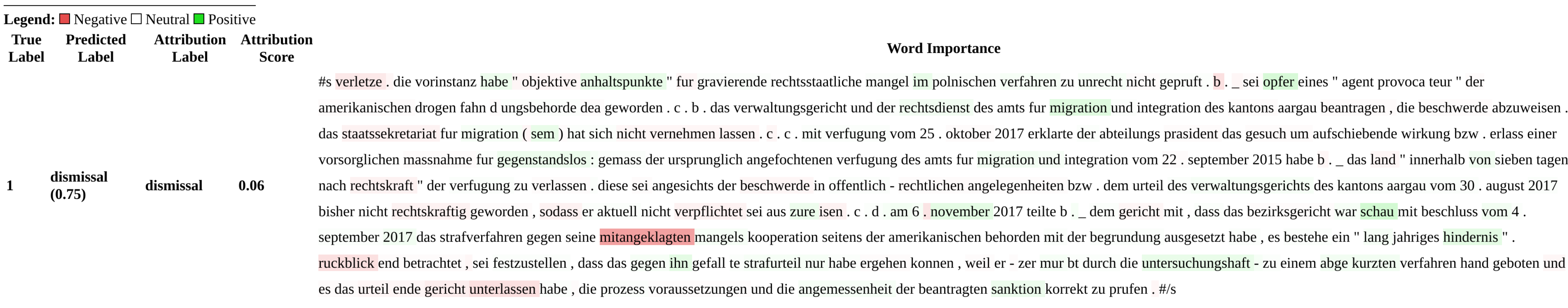}
    \caption{Attention Visualization, Chunk of a Fact, Descriptor: ``Opfer''}
  \label{fig:german_av}
\end{sidewaysfigure}

The 313 summarized facts and 508 chunks were analyzed using attention visualization, selected examples of these attention visualizations are presented in Figures \ref{fig:french_av} and \ref{fig:german_av}. In the French example, the word ``victime'' is shown in green color, indicating a positive attribution towards the ``dismissal'' label, whereas approval (=1) is the true label. We see a similar pattern for the word ``Opfer'' in the German example. For all descriptors, just these two showed a consistent pattern of their attribution being biased towards dismissed, regardless of the final model prediction.  Therefore, we can conclude that our model may tend to make biased predictions for facts containing the aforementioned descriptors.

\clearpage

\subsection{Limitations}

One limitation of this study is the potential validity issues arising from translating descriptors from English into other languages. This is particularly concerning in different cultural contexts where certain words considered “dispreferred” in English may not carry the same connotations in Switzerland or other countries with the same languages. This discrepancy could introduce bias and affect the comparability of the results. Another limitation is the use of descriptors without considering the legal context. This could lead to misjudgments, especially in legal cases often involving victim-perpetrator situations. For instance, descriptors related to victims might be assessed neutrally without context. Similarly, the descriptor “entitled”, translated to “berechtigt” in German, can be perceived neutrally, unlike its potentially negative connotation in English. Ignoring the legal context could bias the results, necessitating critical reflection when interpreting them. Additionally, the chunking process included a decision mechanism that, in case of a tie, favored the majority class “dismissal.” While this method prefers the majority decision, incorporating the classifier's confidence level could provide a more balanced final decision. Moreover, the 512-token limit of the BERT-based model presented a significant challenge. During extractive summarization of facts, this limit led to substantial information loss, as over 50\% of the facts considered exceeded 512 tokens. Consequently, the text contents, including the investigated descriptors, were reduced. In the chunking process, facts were divided into chunks of 512 tokens and classified separately, aiming to consider the entire text content without loss. However, this method did not allow for a complete consideration of the entire context of a fact during classification, potentially resulting in missing contextual information and inaccurate or biased classifications.

\section{Conclusion}\label{sec:conclude}

This work analyzed bias elements in the Swiss Judgement Prediction (SJP) Dataset. Initially, “dispreferred” labeled bias descriptors were extracted from the Holistic Bias Dataset. These descriptors were expanded with synonyms, plural, and gender forms, and made available in German, French, and Italian for the SJP Dataset. Two methods, extractive summarization and splitting long text into chunks, were applied to circumvent the 512-token limit of the BERT-based model, allowing the inclusion of descriptors in the analysis. Potential bias sources were identified using the binomial test to analyze the co-occurrence of these derived descriptors with the SJP Dataset labels. The implementation of the BERT-based model was enhanced by fine-tuning and addressing the imbalanced SJP Dataset using custom trainers to adjust class weights. Finally, attention visualization was used to analyze word attributions of the “dispreferred” descriptors in predicted facts and chunks, examining if these descriptors could impact model performance. Instead of finding social bias, we observed language artifacts for the translations of of the descriptor ``victim'', which is neutral in a legal context. Future work could involve using Large Language Models to identify bias of several types independently from descriptors. Also, this work can be extended by training a model specifically to identify bias, using a BERT-based sequence classification model. Employing a pre-trained model with a token limit over 512 tokens would also be beneficial.  We plan to deepen our insights by extracting legal parties to understand whether the bias affected winning the legal case.


\bibliography{references}

\end{document}